\documentclass[letterpaper,10pt]{IEEEtran}
\usepackage{graphicx,times,amsmath,amssymb,cite,hyperref,subfigure,url}
\usepackage[flushleft]{threeparttable}
\usepackage[lined,ruled,commentsnumbered]{algorithm2e}
 \allowdisplaybreaks

\begin{document}

\title{Pool-Based Sequential Active Learning for Regression}
\author{Dongrui Wu\\
DataNova LLC, USA \\
Email: drwu09@gmail.com.}

\maketitle

\begin{abstract}
Active learning is a machine learning approach for reducing the data labeling effort. Given a pool of unlabeled samples, it tries to select the most useful ones to label so that a model built from them can achieve the best possible performance. This paper focuses on pool-based sequential active learning for regression (ALR). We first propose three essential criteria that an ALR approach should consider in selecting the most useful unlabeled samples: informativeness, representativeness, and diversity, and compare four existing ALR approaches against them. We then propose a new ALR approach using passive sampling, which considers both the representativeness and the diversity in both the initialization and subsequent iterations. Remarkably, this approach can also be integrated with other existing ALR approaches in the literature to further improve the performance. Extensive experiments on 11 UCI, CMU StatLib, and UFL Media Core datasets from various domains verified the effectiveness of our proposed ALR approaches.
\end{abstract}


\section{Introduction}

Active learning (AL) \cite{Settles2009}, a subfield of machine learning, considers the following problem: if the learning algorithm can choose the training data, then which training samples should it choose to maximize the learning performance, under a fixed budget, e.g., the maximum number of labeled training samples? As an example, consider emotion estimation in affective computing \cite{Picard1997}. Emotions can be represented as continuous numbers in the 2D space of arousal and valence \cite{Russell1980}, or in the 3D space of arousal, valence, and dominance \cite{Mehrabian1980}. However, emotions are very subjective, subtle, and uncertain. So, usually multiple human assessors are needed to obtain the groundtruth emotion values for each affective sample (video, audio, image, physiological signal, etc). For example, 14-16 assessors were used to evaluate each video clip in the DEAP dataset \cite{Koelstra2012}, six to 17 assessors for each utterance in the VAM
(\emph{Vera am Mittag} in German, \emph{Vera at Noon} in English) spontaneous speech corpus \cite{Grimm2008}, and at least 110 assessors for each sound in the IADS-2 (International Affective Digitized Sounds
2nd Edition) dataset \cite{Bradley2007}. This is very time-consuming and labor-intensive. How should we optimally select the affective samples to label so that an accurate regression model can be built with the minimum cost (i.e., the minimum number of labeled samples)? That's the typical type of problems that AL targets at.

Many AL approaches have been proposed in the literature \cite{Abe1998,Seung1992,Freund1997,Settles2009,Burbidge2007,Cohn1996,Krogh1995,Cai2017,RayChaudhuri1995,
Demir2014,Settles2008b,Settles2008,Cai2014}. According to the query scenario, they can be categorized into two groups \cite{Sugiyama2009}: \emph{population-based} and \emph{pool-based}. In population-based AL, the test input distribution is known, and training input samples at any desired locations can be queried. Its goal is to find the optimal training input density to generate the training input samples. In pool-based AL, a pool of unlabeled samples is given, and the goal is to optimally choose some to label, so that a model trained from them can best label the remaining samples.

Regardless of whether it is population-based or pool-based, typically AL is iterative \cite{Cai2017}. It first builds a base model from a small number of labeled training samples, and then chooses a few most helpful unlabeled samples and queries for their labels. The newly labeled samples are then added to the training dataset and used to update the model. This process iterates until a termination criterion is met, e.g., the maximum number of iterations, the maximum number of labeled samples, or the desired cross-validation accuracy, is reached. Depending on the number of unlabeled samples to query in each iteration, AL approaches can also be categorized into two types \cite{Cai2017}: \emph{sequential} AL, where one sample is queried each time, and \emph{batch-mode} AL, where multiple samples are queried in each iteration.

This paper focuses on pool-based sequential active learning for regression (ALR). Although numerous AL approaches have been proposed in the literature \cite{Settles2009}, most of them are for classification problems. Among those limited number of ALR approaches \cite{Cai2017,Cai2013,drwuEBMAL2016,Burbidge2007,Sugiyama2009,Yu2010,Willett2006,Sugiyama2006,Cohn1996a,
Freund1997,MacKay1992}, only a few can be used for pool-based sequential ALR \cite{Burbidge2007,Cai2013,Yu2010,drwuEBMAL2016}. In this paper we will review them, point out their limitations, and propose approaches to enhance their performance.

The main contributions of this paper are:
\begin{enumerate}
\item We extend three criteria for AL -- informativeness, representativeness, and diversity -- from classification to regression, and propose a generic framework that can be used to enhance a baseline ALR approach.
\item We instantiate several ALR approaches that consider informativeness, representativeness, and diversity simultaneously, and demonstrate their promising performances in extensive application domains.
\end{enumerate}

The remainder of this paper is organized as follows: Section~\ref{sect:three} introduces three essential criteria that should be considered in ALR, and then compares several existing pool-based sequential ALR approaches against them. Section~\ref{sect:ESAL} proposes several new pool-based sequential ALR approaches. Section~\ref{sect:experiment} describes the datasets to evaluate the effectiveness of the proposed ALR approaches, and the corresponding experimental results. Finally, Section~\ref{sect:conclusions} draws conclusions.

\section{Existing Pool-Based Sequential ALR Approaches} \label{sect:three}

In this section we propose three essential criteria for selecting unlabeled samples in pool-based sequential ALR, and then introduce a few existing pool-based sequential ALR approaches. We also compare these ALR approaches against the three criteria and point out their limitations.

Without loss of generality, we assume the pool consists of $N$ $d$-dimensional samples $\{\mathbf{x}_n\}_{n=1}^N$, $\mathbf{x}_n\in \mathbb{R}^d$, and the first $M_0$ samples have already been labeled with labels $\{y_n\}_{n=1}^{M_0}$.

\subsection{Three Essential Criteria in ALR}

\begin{figure}[htpb]\centering
\includegraphics[width=.8\linewidth,clip]{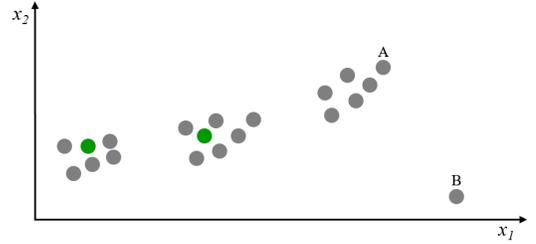}
\caption{Illustration of representativeness and diversity in pool-based sequential ALR.} \label{fig:AL3}
\end{figure}

We propose the following three criteria that should be considered in pool-based sequential ALR for selecting the most useful unlabeled sample to label:
\begin{enumerate}
\item \emph{Informativeness}, which means that the selected sample must contain rich information, so labeling it would significantly benefit the objective function. Informativeness could be measured by uncertainty (entropy, distance to the decision boundary, confidence of the prediction, etc.), expected model change, expected error reduction, and so on \cite{Settles2009}. For example, in query-by-committee (QBC), a popular AL approach for both classification and regression \cite{RayChaudhuri1995}, the informativeness of an unlabeled sample could be computed as the disagreement among the committee members: the more disagreement is, the more uncertain the sample is, and hence the more informative it is.

\item \emph{Representativeness}, which can be evaluated by the number of samples that are similar or close to a target sample (or its density \cite{Settles2009}): the larger the number is, the more representative the target sample is. Clearly, the target sample should not be an outlier. For example, in Fig.~\ref{fig:AL3}, assume we want to build a regression model to predict the output from $x_1$ and $x_2$. The gray circle ``B" is very likely to be an outlier because it is very far away from all other samples in the input space, so labeling it could mislead the regression model and result in overall worse prediction performance. In other words, a sample like ``B" should not be selected for labeling by ALR.

\item \emph{Diversity}, which means that the selected samples should scatter across the full input space, instead of concentrating in a small local region. For example, in Fig.~\ref{fig:AL3} the unlabeled samples form three clusters in the input space, so we should select samples from all three clusters to label, instead of focusing on only one or two of them. Assume the two green circles have been selected and labeled. Then, selecting next a sample from the third cluster (the one contains ``A") seems very reasonable.
\end{enumerate}

We should point out that similar criteria have been used in AL for classification (ALC). For example, Shen et al. \cite{Shen2004a} proposed two multi-criteria-based batch-mode ALC strategies, both of which considered informativeness, representativeness and diversity simultaneously. Their Strategy 1 first chooses a few most informative samples, clusters them, and then selects the cluster centroids for labeling. Their Strategy 2 first computes a score for each sample as a linear combination of its informativeness and representativeness, selects samples with high scores, and further down-selects among them the most diverse ones for labeling. Both strategies are specific to the support vector machine classifier. He et al. \cite{He2014} considered uncertainty, representativeness, information content, and diversity in batch-mode ALC. Let $k$ be the batch size. They compute the information content of an unlabeled sample as uncertainty$\times$representativeness, select the most informative samples, cluster them into $k$ clusters by kernel $k$-means clustering, and finally select the $k$ cluster centers for labeling.

However, to our knowledge, similar ideas have not been explored in ALR, except our recent work on enhanced batch-mode ALR (EBMALR) \cite{drwuEBMAL2016}. And it is not trivial to extend these concepts from classification to regression, because there could be many different strategies to integrate these three criteria, and different strategies could result in significantly different performances. EBMALR \cite{drwuEBMAL2016} is one of such strategies. However, although our previous research \cite{drwuEBMAL2016} showed that it achieved promising performance in \emph{batch-mode} ALR in brain-computer interface, this paper (Section~\ref{sect:results}) shows that it does not perform well in \emph{sequential} ALR. So, how to integrate informativeness, representativeness and diversity to design high-performance pool-based \emph{sequential} ALR is still an open problem.

Next we will introduce several existing pool-based sequential ALR approaches, and compare their rationale against our three criteria.

\subsection{Query-by-Committee (QBC)} \label{sect:QBC}

QBC is a very popular pool-based AL approach for both classification \cite{Abe1998,Seung1992,Freund1997,Settles2009,drwuRSVP2016} and regression \cite{Burbidge2007,Cohn1996,Krogh1995,RayChaudhuri1995,Settles2009,Demir2014}. Its basic idea is to build a committee of learners from existing labeled training dataset (usually through bootstrapping and/or different learning algorithms), and then select from the pool the unlabeled samples on which the committee disagrees the most to label.

In this paper we use the pool-based QBC for regression approach proposed by RayChaudhuri and Hamey \cite{RayChaudhuri1995}. It first bootstraps the $M_0$ labeled samples into $P$ copies, each containing $M_0$ samples but with duplicates, and builds a regression model from each copy, i.e., the committee consists of $P$ regression models. Let the $p$th model's prediction for the $n$th unlabeled sample be $y_n^p$. Then, for each of the $N-M_0$ unlabeled sample, it computes the variance of the $P$ individual predictions, i.e.,
\begin{align}
\sigma_n=\frac{1}{P}\sum_{p=1}^P\left(y_n^p-\bar{y}_n\right)^2, \quad n=M_0+1,...,N
\end{align}
where $\bar{y}_n=\frac{1}{P}\sum_{p=1}^P y_n^p$, and then selects the sample with the maximum variance to label.

Comparing against the three criteria for ALR, QBC only considers the informativeness, but not the representativeness and the diversity.

\subsection{Expected Model Change Maximization (EMCM)} \label{sect:EMCM}

Expected model change maximization (EMCM) is also a very popular AL approach for classification \cite{Settles2009,Settles2008b,Settles2008,Cai2014}, regression \cite{Cai2013,Cai2017}, and ranking \cite{Donmez2008}. Cai et al. \cite{Cai2013} proposed an EMCM approach for both linear and nonlinear regression. In this subsection we introduce their linear approach, as only linear regression is considered in this paper.

EMCM first uses all $M_0$ labeled samples to build a linear regression model. Let its prediction for the $n$th sample $\mathbf{x}_n$ be $\hat{y}_n$. Then, like in QBC, EMCM also uses bootstrap to construct $P$ linear regression models. Let the $p$th model's prediction for the $n$th sample $\mathbf{x}_n$ be $y_n^p$. Then, for each of the $N-M_0$ unlabeled samples, it computes
\begin{align}
g(\mathbf{x}_n)=\frac{1}{P}\sum_{p=1}^P\left\| (y_n^p-\hat{y}_n)\mathbf{x}_n\right\|, \quad n=M_0+1,...,N
\end{align}
EMCM selects the sample with the maximum $g(\mathbf{x}_n)$ to label.

Comparing against the three criteria for ALR, EMCM only considers the informativeness, but not the representativeness and the diversity.

\subsection{Greedy Sampling (GS)} \label{sect:GS}

Yu and Kim \cite{Yu2010} proposed several very interesting passive sampling techniques for regression. Instead of finding the most informative sample based on the learned regression model, as in QBC and EMCM, they select the sample based on its geometric characteristics in the feature space. An advantage of passive sampling is that it does not require updating the regression model and evaluating the unlabeled samples in each iteration, so it is independent of the regression model.

In this paper we use the greedy sampling (GS) for regression approach \cite{Yu2010}, which is easy to implement, and showed promising performance in \cite{Yu2010}. Its basic idea is to select a new sample in a greedy way such that it is located far away from the previously selected and labeled samples. More specifically, for each of the $N-M_0$ unlabeled sample $\{\mathbf{x}_n\}_{n=M_0+1}^N$, it computes its distance to each of the $M_0$ labeled samples, i.e.,
\begin{align}
d_{nm}=||\mathbf{x}_n-\mathbf{x}_m||,\quad m=1,...,M_0; n=M_0+1,...,N
\end{align}
then it computes $\underline{d}_n$ as the minimum distance from $\mathbf{x}_n$ to the $M_0$ labeled samples, i.e.,
\begin{align}
\underline{d}_n=\min_m d_{nm},\quad n=M_0+1,...,N
\end{align}
and selects the sample with the maximum $\underline{d}_n$ to label.

Comparing against the three criteria for ALR, GS only considers the diversity, but not the informativeness and the representativeness.

\subsection{Enhanced Batch-Mode ALR (EBMALR)} \label{sect:EBMALR}

We have already seen that each of the above three ALR approaches only considers one of the three essential criteria for ALR, so there is room for improvement. Additionally, all of them assume that we already have $M_0$ initially labeled samples for training. Usually these $M_0$ samples are randomly selected, because the regression models cannot be constructed at the very beginning when no or very few labeled samples are available (and hence QBC and EMCM cannot be applied). However, there can still be better initialization approaches to select more representative and diverse seedling samples, without using any label information. One such approach, EBMALR \cite{drwuEBMAL2016}, was proposed recently to consider simultaneously informativeness, representativeness and diversity to enhance QBC and EMCM. Theoretically, batch-mode ALR can also be used for sequential ALR, by setting the batch size to one. Algorithm~\ref{alg:EAL} shows the EBMALR algorithm when the batch size is one. It first uses $k$-means clustering to initialize $d$ samples that are representative and diverse, and then uses a baseline ALR approach, such as QBC or EMCM, to select subsequent samples sequentially.

Compared with QBC, EMCM and GS, EBMALR identifies outliers and excludes them from being selected, and considers both representativeness and diversity in initializing the first $d$ samples. The original EBMALR (when the batch size is larger than one) considers both diversity and informativeness in each subsequent iteration, but when the batch size becomes one, EBMALR is no longer able to consider the diversity among the selected samples. As a result, its performance degrades significantly, as will be demonstrated in Section~\ref{sect:results}.

\begin{algorithm}[h] %\DontPrintSemicolon
\KwIn{$N$ unlabeled samples, $\{\mathbf{x}_n\}_{n=1}^N$, where $\mathbf{x}_n\in \mathbb{R}^d$\;
\hspace*{9mm} $M$, the maximum number of labeled samples to query\;
\hspace*{9mm} $\gamma$, the threshold for outlier identification}
\KwOut{The regression model $f(\mathbf{x})$.}
\tcp{Identify the outliers}
$S=\{\mathbf{x}_n\}_{n=1}^N$\;
$hasOutliers$=True\;
\While{$hasOutliers$}{
Perform $k$-means clustering on $S$ to obtain $d$ clusters, $C_i$, $i=1,...,d$\;
Set $p_i=|C_i|$\;
$hasOutliers$=False\;
\For{$i=1,...,k$}{
\If{$p_i\le \max(1,\gamma N)$}{
$S=S\setminus C_i$\;
$hasOutliers$=True\;}}}
\tcp{Initialize $d$ labeled samples}
\For{$i=1,...,d$}{
Select the sample closest to the centroid of $C_i$ to label\;}
\tcp{End initialization}
\For{$m=d+1,...,M$}{
{Perform a baseline ALR (e.g., QBC or EMCM) on $S$ to select a sample for labeling\;}}
Construct the regression model $f(\mathbf{x})$ from the $M$ labeled samples.
\caption{The EBMALR algorithm, when the batch size is 1.} \label{alg:EAL}
\end{algorithm}

\subsection{Design of Experiments (DOE)}

Design of experiments (DOE) has been widely studies in statistics and used in various industries, for ``\emph{exploring new processes and gaining increased knowledge of existing processes, followed by optimising these processes for achieving world-class performance \cite{Antony2014}.}" Its primary goal is usually to extract the maximum amount of information from as few observations as possible, which is very similar to ALR. There are two typical categories of DOEs \cite{Antony2014}:
\begin{enumerate}
\item \emph{Screening designs}, which are smaller experiments to identify the critical few factors from the many potential trivial factors.
\item \emph{Optimal designs}, which are larger experiments that investigate interactions of terms and nonlinear responses, and are conducted at more than two levels for each factor.
\end{enumerate}
Optimal designs are particularly relevant to ALR. They provide theoretical criteria to choosing a set of points to label, for a specific set of assumptions and objectives. Compared with optimal designs, ALR approaches are generally more heuristic. In this paper we only consider ALR approaches.

\section{Our Proposed ALR Approaches} \label{sect:ESAL}

In this section we propose first a basic pool-based sequential ALR approach by considering simultaneously representativeness and diversity (RD), and then strategies to integrate it with QBC, EMCM and GS to further improve the performance.

\subsection{The Basic RD ALR Approach} \label{sect:RD}

Assume initially none of the $N$ samples in the pool is labeled. Our proposed basic RD ALR approach consists of two parts: 1) Better initialization of the first a few samples by considering both representativeness and diversity; and, 2) Generating a new sample in each subsequent iteration by also considering both representativeness and diversity.

Since the input space has $d$ dimensions, it is desirable to have at least $d$ initially labeled samples to construct a reasonable linear regression model\footnote{It is also possible to initialize fewer than $d$ samples to construct a linear regression model, by using regularized regression such as ridge regression and LASSO. However, here we assume $d$ is small, and initialize $d$ samples directly for simplicity.}. To find the optimal locations of these $d$ samples, we perform $k$-means ($k=d$) clustering on the $N$ unlabeled samples, and then select from each cluster the sample closest to the cluster centroid for labeling. This initialization ensures representativeness, because each sample is a good representation of the cluster it belongs to. It also ensures diversity, because these $d$ clusters cover the full input space of $\mathbf{x}$.

The idea of using clustering for sample selection in ALR was motivated by similar ideas in ALC. For example, Nguyen and Smeulders \cite{Nguyen2004} used $k$-medoids clustering to select representative and diverse samples. Kang, Ryu and Kwon \cite{Kang2004} used $k$-means clustering to partition the unlabeled samples into different clusters, and then selected from each cluster the sample closest to its centroid as the most representative one. Hu, Namee and Delany \cite{Hu2010b} used deterministic clustering methods (furthest-first-traversal, agglomerative hierarchical clustering, and affinity propagation clustering) to initialize the samples, to avoid variations introduced by non-deterministic clustering approaches such as $k$-medoids and $k$-means. Krempl, Ha and Spiliopoulou \cite{Krempl2015} proposed a clustering-based optimized probabilistic active learning approach for online streaming ALC. However, to our knowledge, there have not yet existed any pool-based sequential ALR approaches that use clustering to initialize the samples and also perform subsequent selections.

After the first $d$ samples are initialized by considering both representativeness and diversity, next we start the iterative ALR process, where a new sample is selected for labeling in each iteration. Consider the first iteration, where we already have $d$ labeled samples, and need to determine which sample from the $N-d$ unlabeled ones should be further selected for labeling. In the basic RD algorithm, we first perform $k$-means clustering on all $N$ samples, where $k=d+1$. Since there are only $d$ labeled samples, at least one cluster does not contain any labeled sample. In practice some clusters may contain multiple labeled samples, so usually there are more than one clusters that do not contain any labeled sample. We then identify the largest cluster that does not contain any labeled sample as the current most representative cluster, and select the sample closest to its centroid for labeling. Note that this selection strategy also ensures diversity, because the identified cluster locates differently from all other clusters that already contain labeled samples. We then repeat this process to generate more labeled samples, until the maximum number of labeled samples is reached.

The pseudo-code of the basic RD ALR approach is given in Algorithm~\ref{alg:RDAL}, where \emph{Option 1} is used. Similar to GS, the basic RD approach also uses passive sampling, which does not require updating the regression model and evaluating the unlabeled sample in each iteration. So, it is independent of the regression model.

\begin{algorithm}[h] %\DontPrintSemicolon
\KwIn{$N$ unlabeled samples, $\{\mathbf{x}_n\}_{n=1}^N$, where $\mathbf{x}_n\in \mathbb{R}^d$\;
\hspace*{9mm} $M$, the maximum number of labeled samples to query}
\KwOut{The regression model $f(\mathbf{x})$.}
\tcp{Initialize $d$ labeled samples}
Perform $k$-means clustering on $\{\mathbf{x}_n\}_{n=1}^N$, where $k=d$\;
Select from each cluster the sample closest to its centroid, and query for its label\;
\tcp{End initialization}
\For{$m=d+1,...,M$}{
{Perform $k$-means clustering on $\{\mathbf{x}_n\}_{n=1}^N$, where $k=m$\;
Identify the largest cluster that does not already contain a labeled sample\;
\emph{Option 1}: Select the sample closest to the cluster centroid for labeling\;
\emph{Option 2}: Use QBC (Section~\ref{sect:QBC}) to select a sample from the cluster for labeling\;
\emph{Option 3}: Use EMCM (Section~\ref{sect:EMCM}) to select a sample from the cluster for labeling\;
\emph{Option 4}: Use GS (Section~\ref{sect:GS}) to select a sample from the cluster for labeling\; }}
Construct the regression model $f(\mathbf{x})$ from the $M$ labeled samples.
\caption{The proposed RD ALR algorithm, and its variations.} \label{alg:RDAL}
\end{algorithm}

\subsection{Integrate RD with QBC, EMCM, and GS} \label{sect:IRD}

Interestingly, the basic RD ALR approach can be easily integrated with an existing pool-based sequential ALR approach for better performance. The pseudo-code is also shown in Algorithm~\ref{alg:RDAL}, where \emph{Option} 2 or 3 or 4 is used. The initialization is the same as the basic RD ALR approach. In each iteration, it also selects a sample from the largest cluster that does not already contain a labeled sample for labeling. However, instead of selecting the one closest to its centroid, as in the basic RD ALR approach, now it uses QBC or EMCM or GS to select the most informative or most diverse sample to label. We expect that when QBC or EMCM is used, the integrated RD ALR approach can achieve better performance than the basic RD ALR approach, because now informativeness, representativeness and diversity are considered simultaneously.

\subsection{Differences from EBMALR}

Our proposed ALR approaches have some similarity with EBMALR \cite{drwuEBMAL2016}, e.g., clustering is used to ensure representativeness and diversity. However, there are several significant differences:
\begin{enumerate}
\item This paper considers pool-based \emph{sequential} ALR, whereas \cite{drwuEBMAL2016} considered pool-based \emph{batch-model} ALR. Theoretically, sequential ALR can be viewed as a special case of batch-model ALR, when the batch size equals one. However, as pointed out in Section~\ref{sect:EBMALR}, when the batch size becomes one, EBMALR is no longer able to consider the diversity among the selected samples. As a result, its performance becomes significantly worse than the proposed approaches in this paper, as will be shown in Section~\ref{sect:results}.
\item This paper explicitly defines informativeness, representativeness and diversity as three criteria that should be considered in ALR, whereas EBMALR did not (although it implicitly used these concepts).
\item EBMALR considered also how to exclude outliers from being selected, but it required a user-defined parameter. Through extensive experiments, we found that this part is not critical in most applications, so this paper does not include it. As a result, our new algorithms do not require any user-defined hyper-parameters, which are easier to use.
\item In each subsequent iteration, EBMALR (when the batch size is larger than one) considered first the informativeness and then the diversity, but this paper considers first the diversity and then the informativeness or representativeness. Experiments showed that the latter results in better performances.
\item This paper introduces a greedy sampling approach (Section~\ref{sect:GS}) for ALR, and also proposes a new RD approach (Section~\ref{sect:RD}), both of which were not included in \cite{drwuEBMAL2016}.
\item This paper compares the performances of nine ALR approaches on 11 datasets from various domains, whereas \cite{drwuEBMAL2016} only compared five ALR approaches in a brain-computer interface application.
\end{enumerate}

\section{Experiments and Results} \label{sect:experiment}

Extensive experiments are performed in this section to demonstrate the performance of the basic and integrated RD ALR approaches.

\subsection{Datasets}

We used 10 datasets from the UCI Machine Learning Repository\footnote{\url{http://archive.ics.uci.edu/ml/index.php}} and the CMU StatLib Datasets Archive\footnote{\url{http://lib.stat.cmu.edu/datasets/}} that have been used in previous ALR experiments \cite{Cai2013,Cai2017,Yu2010}. We also used an IADS-2 dataset on affective computing from the University of Florida Media Core\footnote{\url{http://csea.phhp.ufl.edu/media.html\#midmedia}}. It consists of 167 acoustic emotional stimuli for experimental investigations of emotion and attention. 76 acoustic features were extracted \cite{drwuACII2018}, and principle component analysis was used to reduce them to 10 features. The goal was to estimate the continuous arousal value from these 10 features. The summary of these datasets is given in Table~\ref{tab:datasets}. They cover a large variation of application domains.

\begin{table}[h]
\caption{Summary of the 11 regression datasets.} \label{tab:datasets}
\centering \setlength{\tabcolsep}{.8mm}
\begin{threeparttable}
\begin{tabular}{l|cccccc}   \hline
 & &  No. of         &  No. of    &   No. of   &   No. of  & No. of           \\
Dataset & Source& samples        & raw   &   numerical  &   categorical  & total           \\
& &  &  features  &   features &   features &  features       \\ \hline
Concrete-CS\tnote{1}  &  UCI & 103  &   7 &7 &0 & 7 \\
IADS-Arousal\tnote{2} & UFL & 167 & 10 & 10 & 0 & 10 \\
Yacht\tnote{3}  &  UCI &   308  &   6&6 & 0& 6\\
autoMPG\tnote{4}  & UCI &   392 & 7  &    6     & 1 & 9 \\
NO2\tnote{5}  &  StatLib &   500  &     7 & 7 & 0   &  7  \\
Housing\tnote{6} &  UCI & 506 &  13 & 13 & 0    &  13 \\
CPS\tnote{7}  &   StatLib & 534  &   11 & 8 &3     &  19 \\
Concrete\tnote{8}& UCI & 1030 & 8 & 8 & 0 & 8  \\
Airfoil\tnote{9} & UCI & 1503 & 5 & 5 & 0 & 5 \\
Wine-red\tnote{10} & UCI & 1599  &  11 & 11 & 0    &   11 \\
Wine-white\tnote{10} & UCI & 4898 &  11 & 11 & 0    &   11 \\  \hline
\end{tabular}
  \begin{tablenotes}
\item[1] \url{https://archive.ics.uci.edu/ml/datasets/Concrete+Slump+Test}
\item[2] \url{http://csea.phhp.ufl.edu/media.html#midmedia}
\item[3] \url{https://archive.ics.uci.edu/ml/datasets/Yacht+Hydrodynamics}
\item[4] \url{https://archive.ics.uci.edu/ml/datasets/auto+mpg}
\item[5] \url{http://lib.stat.cmu.edu/datasets/}
\item[6] \url{https://archive.ics.uci.edu/ml/machine-learning-databases/housing/}
\item[7] \url{http://lib.stat.cmu.edu/datasets/CPS_85_Wages}
\item[8] \url{https://archive.ics.uci.edu/ml/datasets/Concrete+Compressive+Strength}
\item[9] \url{https://archive.ics.uci.edu/ml/datasets/Airfoil+Self-Noise}
\item[10] \url{https://archive.ics.uci.edu/ml/datasets/Wine+Quality}
  \end{tablenotes}
\end{threeparttable}
\end{table}

Some datasets contain both numerical and categorical features. For example, the autoMPG dataset contains seven raw features, among which six are numerical and one is categorical (Origin: US, Japan, Germany). We used one-hot encoding to covert the categorical values into numerical values, e.g., Origin-US was encoded as $[1,0,0]$, Origin-Japan $[0,1,0]$, and Origin-Germany $[0,0,1]$. In this way, the converted feature space has $6+3=9$ dimensions. Categorical features in other datasets were converted similarly before regression. We then normalized each dimension of the feature space to mean zero and standard deviation one.

\subsection{Algorithms}

We compared the performances of nine different sample selection strategies:
\begin{enumerate}
\item Baseline (\texttt{BL}), which randomly selects all $M$ samples.
\item \texttt{RD}, which is our basic RD ALR algorithm introduced in Section~\ref{sect:RD}.
\item \texttt{QBC}, which has been introduced in Section~\ref{sect:QBC}. The first $d$ labeled samples are randomly initialized.
\item \texttt{RD-QBC}, which is RD integrated with QBC, introduced in Section~\ref{sect:IRD}.
\item \texttt{EMCM}, which has been introduced in Section~\ref{sect:EMCM}. The first $d$ labeled samples are randomly initialized.
\item \texttt{EEMCM}, which is the EBMALR approach introduced in Algorithm~\ref{alg:EAL}, when EMCM is used as the base ALR approach.
\item \texttt{RD-EMCM}, which is RD integrated with EMCM, introduced in Section~\ref{sect:IRD}.
\item \texttt{GS}, which has been introduced in Section~\ref{sect:GS}. The first $d$ labeled samples are randomly initialized.
\item \texttt{RD-GS}, which is RD integrated with GS, introduced in Section~\ref{sect:IRD}.
\end{enumerate}
All nine approaches built a ridge regression model from the labeled samples, with ridge parameter $\sigma=0.01$. We used ridge regression instead of ordinary linear regression because the number of labeled samples is very small, so ridge regression, with regularization on the coefficients, generally results in better performance than the ordinary linear regression.

\subsection{Evaluation Process} \label{sect:process}

There could be two model evaluation strategies: 1) \emph{inductive learning}, in which we learn a model from labeled samples, and try to evaluate it on samples we have not seen or known about; and, 2) \emph{transductive learning}, in which we try to evaluate the model on a known (test) set of unlabeled examples. Specific to pool-based ALR, the former means labeling a small amount of samples from a fixed pool, building a regression model, and then predicting the outputs of the remaining unlabeled samples in the \emph{same} pool, whereas the latter means labeling a small amount of samples from a fixed pool, building a regression model, and then predicting the outputs of unlabeled samples from \emph{another} pool. This paper mainly focuses on transductive learning, but will also briefly report results on inductive learning in Section~\ref{sect:In} (more results can be found in the Supplementary Materials). Generally they are very similar.

The detailed evaluation process was similar to that used in our previous research on pool-based batch-mode ALR \cite{drwuEBMAL2016}. For each dataset, let $\mathbf{P}$ be the pool of all samples. We first randomly selected 80\% of the total samples as our training pool\footnote{For a fixed pool, \texttt{EEMCM}, \texttt{RD}, \texttt{RD-QBC}, \texttt{RD-EMCM} and \texttt{RD-GS} give a deterministic selection sequence because there is no randomness involved (assume $k$-means clustering always converges to its global optimum). So, we need to vary the pool in order to study the statistical properties of them. We did not use the traditional bootstrap approach, i.e., sampling with replacement to obtain the same number of samples as the original pool, because bootstrap introduces duplicate samples in the new pool, whereas in practice usually a pool does not contain duplicates.} (denoted as $\mathbf{P}_{80}$), initialized the first $d$ labeled samples ($d$ is the number of total features) either randomly or by \texttt{EEMCM}/\texttt{RD}, identified one sample to label in each subsequent iteration by different algorithms, and built a ridge regression model. The maximum number of samples to be labeled, $M$, was 10\% of the size of $\mathbf{P}_{80}$. For datasets too small or too large, we constrained $M\in[20,60]$.

In the inductive learning setting, the model performance was evaluated on the 20\% remaining samples that are in $\mathbf{P}$ but not in $\mathbf{P}_{80}$, whereas in the transductive learning setting, the model performance was evaluated on the samples in $\mathbf{P}_{80}$. We ran the above evaluation process 100 times for each dataset and each algorithm, to obtain statistically meaningful results.

\subsection{Performance Measures} \label{sect:process}

After each iteration of each algorithm, we computed the root mean squared error (RMSE) and correlation coefficient (CC) as the performance measures.

In transductive learning, because different algorithms selected different samples to label, the remaining unlabeled samples in the pool were different for each algorithm, so we cannot compare their performances based on the remaining unlabeled samples. Because in pool-based ALR the goal is to build a regression model to label all samples in the pool as accurately as possible, we computed the RMSE and CC using all samples in the pool, where the labels for the $m$ selected samples were their true labels, and the labels for the remaining $N-m$ unlabeled samples were the predictions from the ridge regression model.

Let $y_n$ be the true label for $\mathbf{x}_n$, and $\hat{y}_n$ be the prediction from the ridge regression model. Without loss of generality, assume the first $m$ samples are selected by an algorithm and hence their true labels are known. Then,
\begin{align}
RMSE&=\left[\frac{1}{N}\sum_{n=1}^N(y_n-y_n')^2\right]^{1/2}\\
CC&=\frac{\sum_{n=1}^N(y_n-\bar{y})(y_n'-\bar{y}')}
{\sqrt{\sum_{n=1}^N(y_n-\bar{y})^2}\sqrt{\sum_{n=1}^N(y_n'-\bar{y}')^2}}
\end{align}
where
\begin{align}
y_n'&=\left\{\begin{array}{ll}
              y_n, & n=1,...,m \\
              \hat{y}_n, & n=m+1,...,N
            \end{array}\right.\\
\bar{y}&=\frac{1}{N}\sum_{n=1}^Ny_n;\quad \bar{y}'=\frac{1}{N}\sum_{n=1}^Ny_n'
\end{align}

Note that we should consider the RMSE as the primary performance measure, because it is directly optimized in the objective function of ridge regression (CC is not). Generally as the RMSE decreases, the CC should increase, but there is no guarantee. In other words, we expect that an ALR approach performing well on the RMSE should also perform well on the CC, but this is not always true. So, the CC can only be viewed as a secondary performance measure.

In inductive learning, the RMSE and CC can be computed directly on the 20\% samples that are in $\mathbf{P}$ but not in $\mathbf{P}_{80}$.

\subsection{Experimental Results} \label{sect:results}

\begin{figure*}[htpb]
\subfigure[]{\label{fig:ConcreteCS}     \includegraphics[width=.32\linewidth,clip]{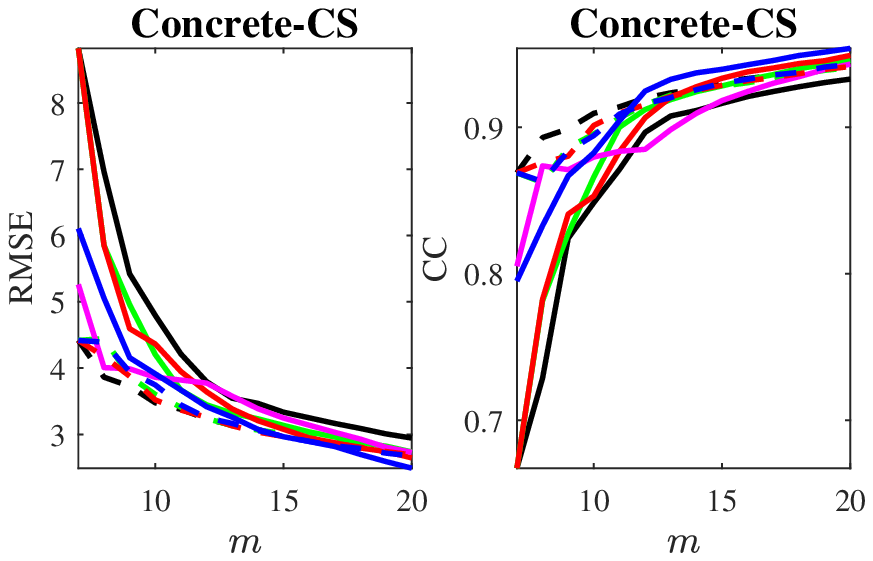}}
\subfigure[]{\label{fig:ConcreteFlow}     \includegraphics[width=.32\linewidth,clip]{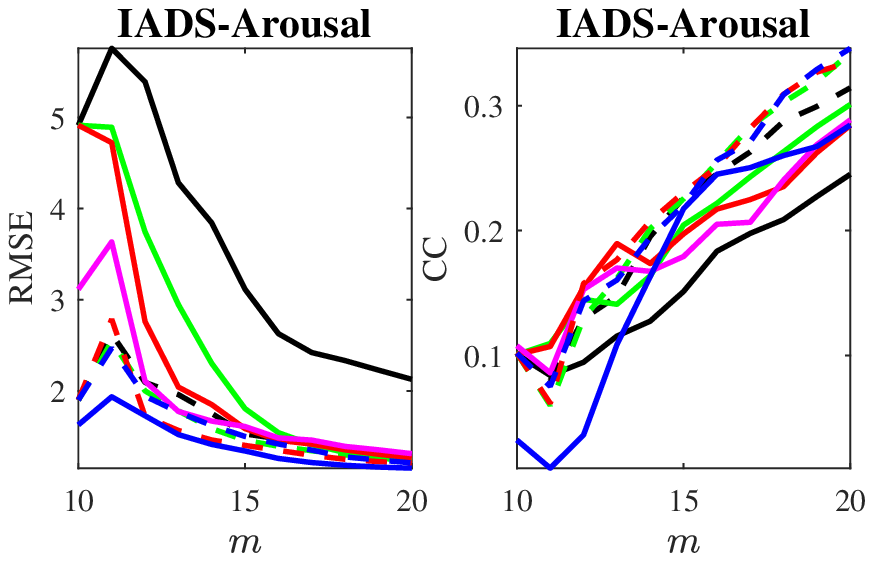}}
\subfigure[]{\label{fig:Yacht}     \includegraphics[width=.32\linewidth,clip]{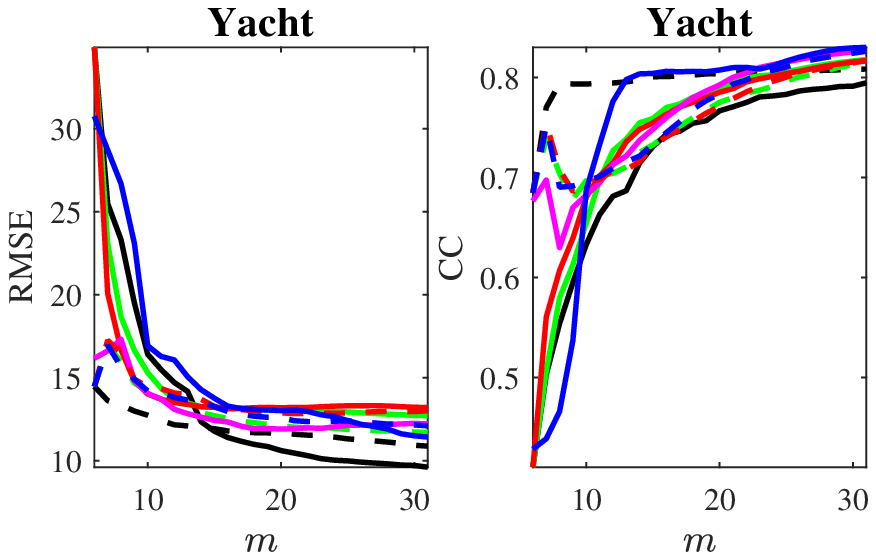}}
\subfigure[]{\label{fig:autoMPG}     \includegraphics[width=.32\linewidth,clip]{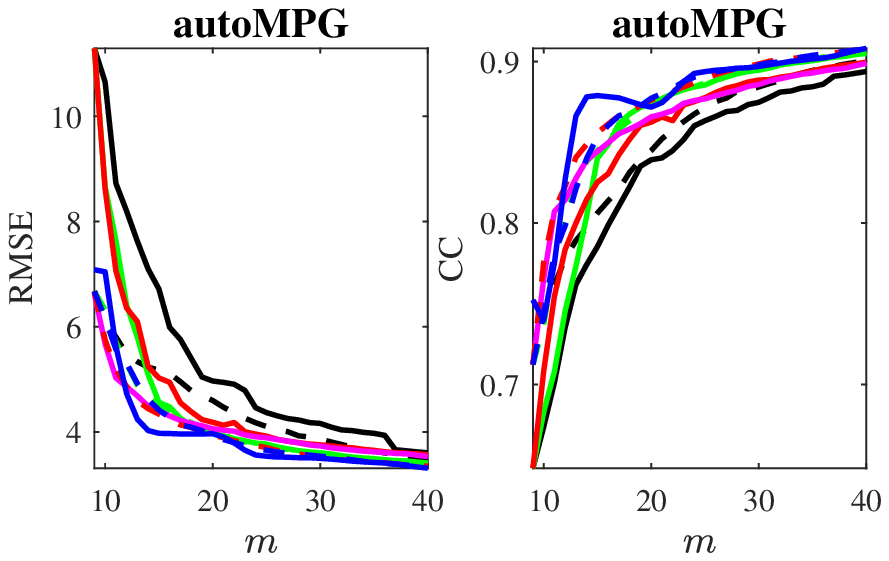}}
\subfigure[]{\label{fig:NO2}     \includegraphics[width=.32\linewidth,clip]{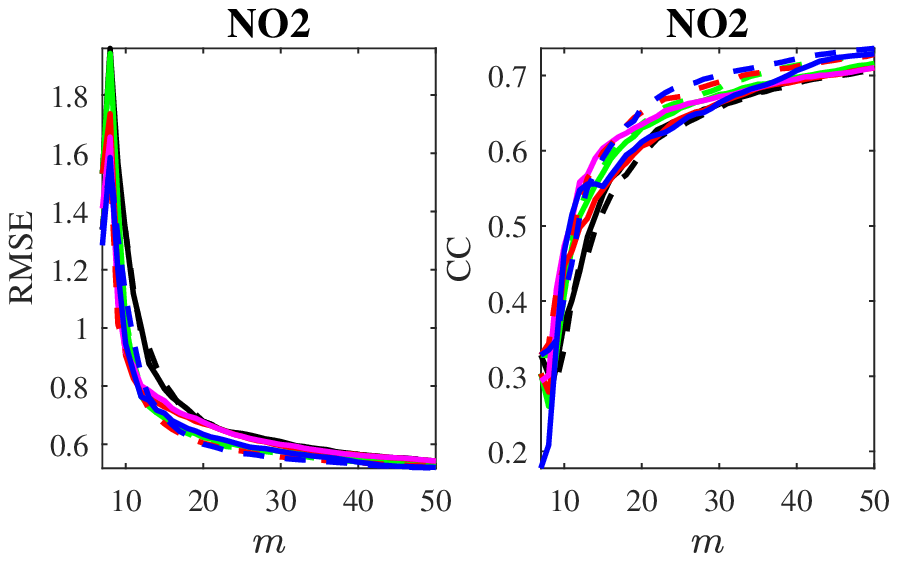}}
\subfigure[]{\label{fig:Housing}     \includegraphics[width=.32\linewidth,clip]{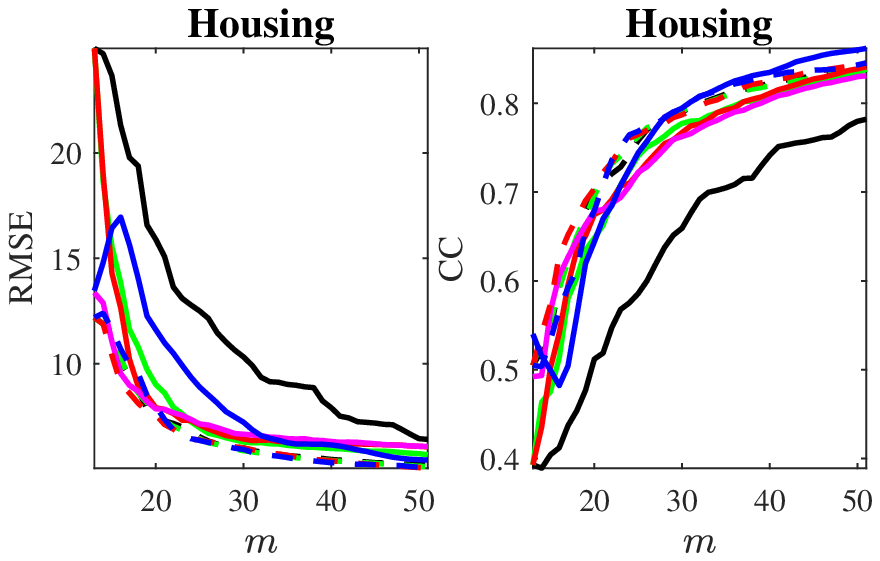}}
\subfigure[]{\label{fig:CPS}     \includegraphics[width=.32\linewidth,clip]{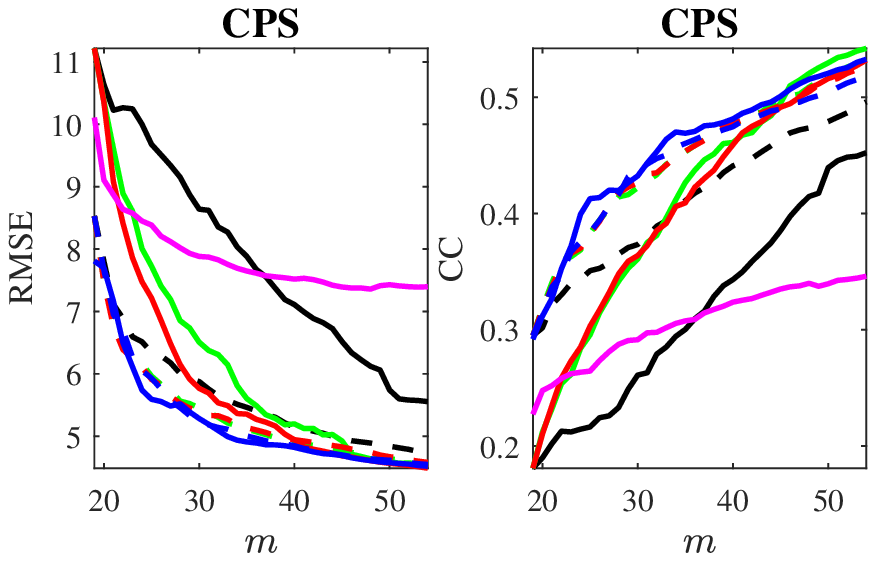}}
\subfigure[]{\label{fig:Concrete}     \includegraphics[width=.32\linewidth,clip]{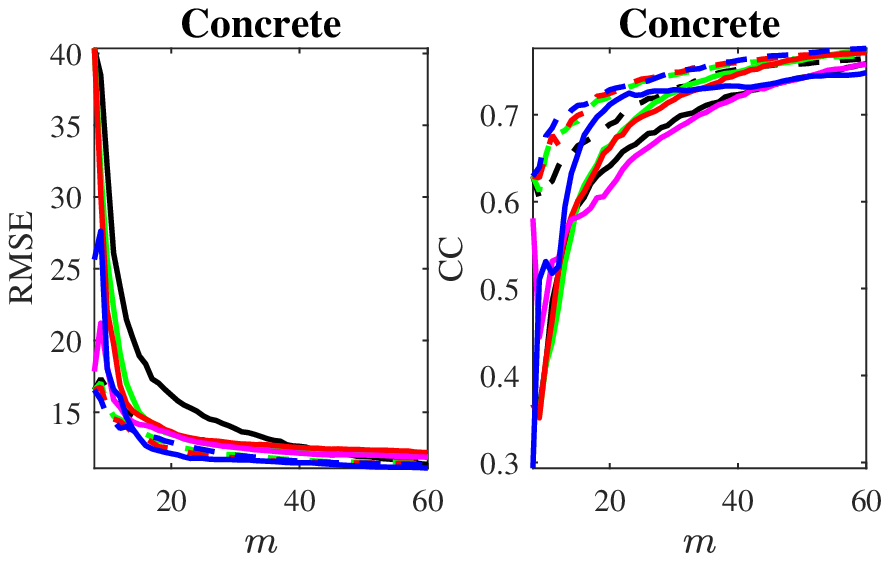}}
\subfigure[]{\label{fig:Airfoil}     \includegraphics[width=.32\linewidth,clip]{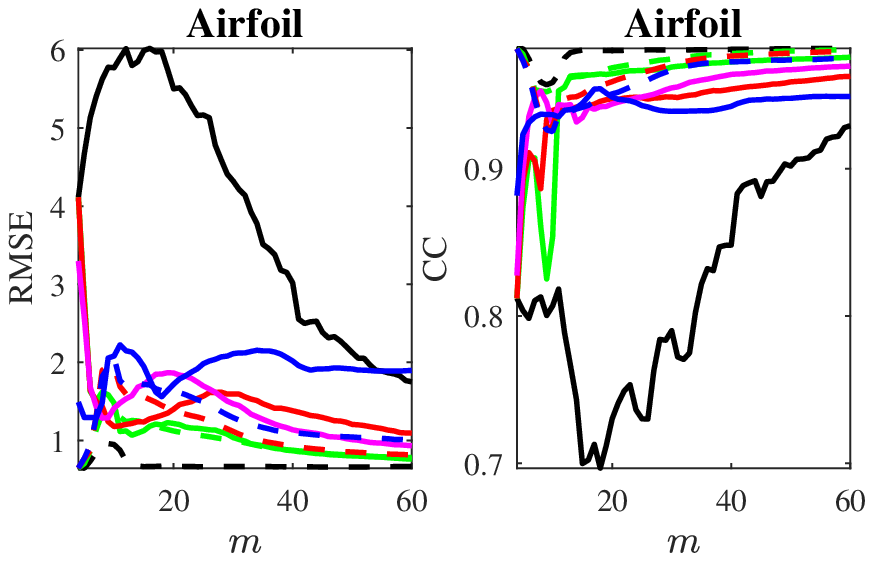}}
\subfigure[]{\label{fig:Wine-red}     \includegraphics[width=.32\linewidth,clip]{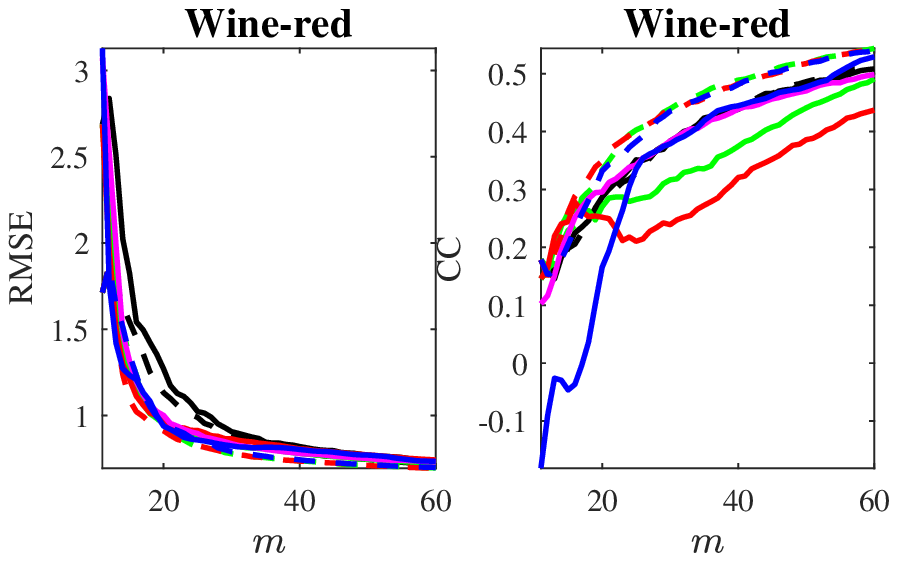}}
\subfigure[]{\label{fig:Wine-white}     \includegraphics[width=.45\linewidth,clip]{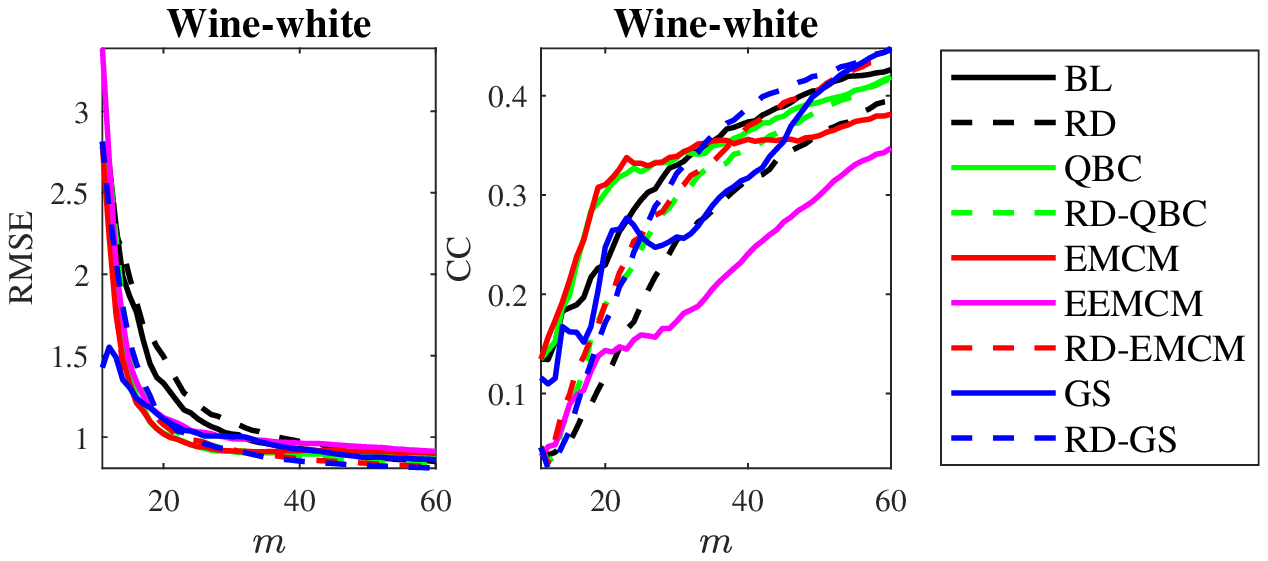}}
\caption{Performances of the nine algorithms on the 11 datasets in transductive learning, averaged over 100 runs. (a) Concrete-CS; (b) IADS-Arousal; (c) Yacht; (d) autoMPG; (e) NO2; (f) Housing; (g) CPS; (h) Concrete; (i) Airfoil; (j) Wine-red; (k) Wine-white.} \label{fig:results11}
\end{figure*}

The RMSEs and CCs for the nine algorithms on the 11 datasets in transductive learning, averaged over 100 runs, are shown in Fig.~\ref{fig:results11}. Observe that:
\begin{enumerate}
\item Generally as $m$ increased, all nine algorithms achieved better performance (smaller RMSE and larger CC), which is intuitive, because more labeled training samples generally result in a more reliable ridge regression model.
\item \texttt{RD}, \texttt{QBC}, \texttt{EMCM}, \texttt{EEMCM} and \texttt{GS} achieved better performances than \texttt{BL} on almost all datasets, suggesting that all these ALR approaches were effective.
\item Generally \texttt{RD-QBC} achieved better performance than both \texttt{RD} and \texttt{QBC}, \texttt{RD-EMCM} achieved better performance than both \texttt{RD} and \texttt{EMCM}, and \texttt{RD-GS} achieved better performance than both \texttt{RD} and \texttt{GS}. These results suggest that our proposed \texttt{RD} ALR approach is complementary to \texttt{QBC}, \texttt{EMCM} and \texttt{GS}, and hence integrating them can outperform each individual ALR approach.
\end{enumerate}

\begin{figure}[htpb]\centering
\subfigure[]{\label{fig:AUC-RMSE}     \includegraphics[width=.88\linewidth,clip]{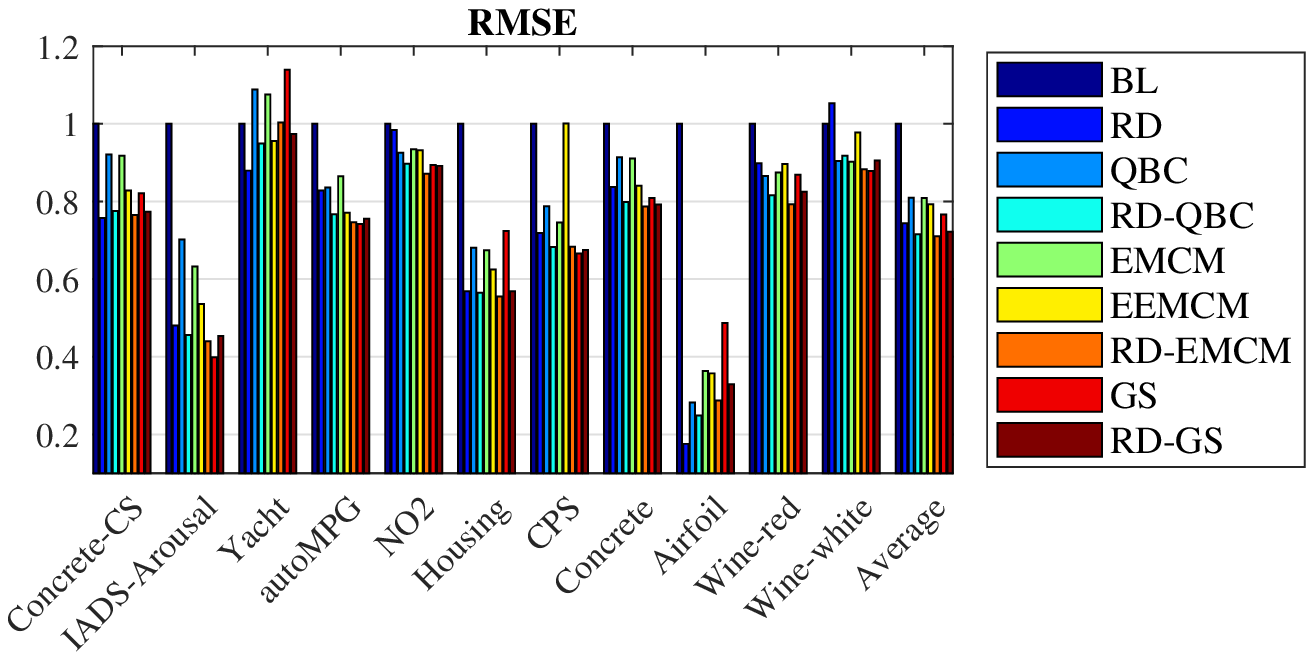}}
\subfigure[]{\label{fig:AUC-CC}     \includegraphics[width=.88\linewidth,clip]{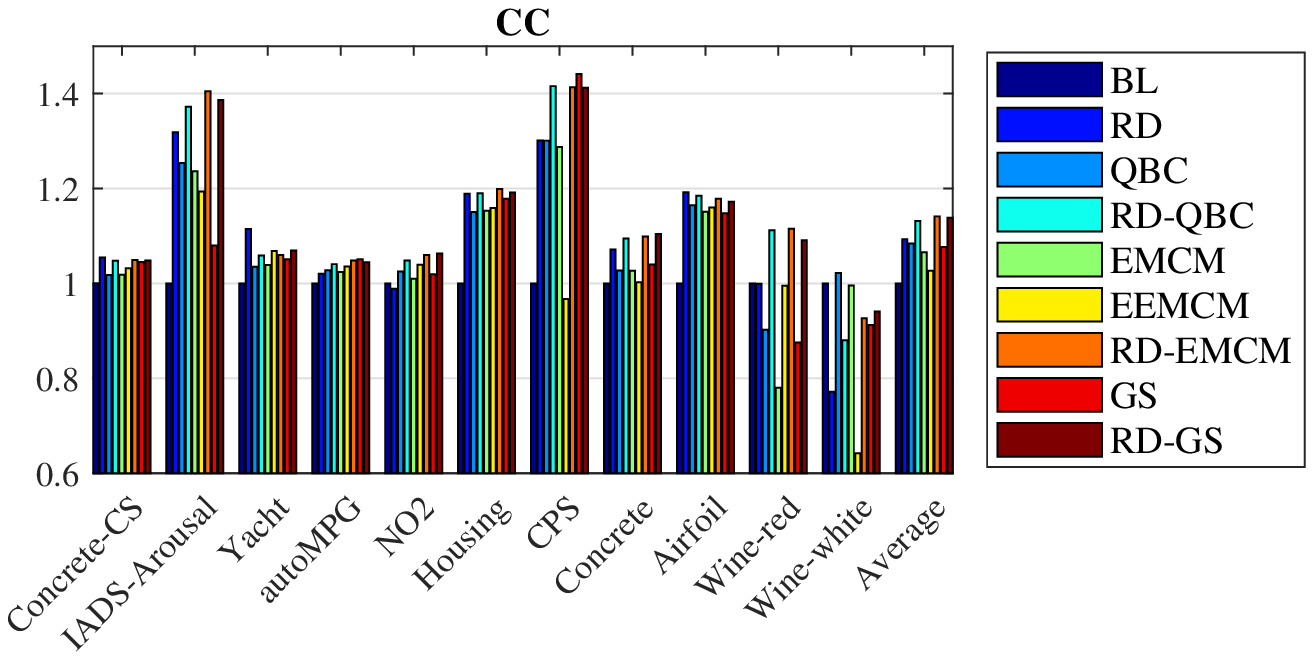}}
\caption{AUCs of the nine algorithms on the 11 datasets in transductive learning. (a) RMSE; (b) CC.} \label{fig:AUC}
\end{figure}

To see the forest for the trees, we also define an aggregated performance measure called the area under the curve (AUC) for the average RMSE and the average CC on each of the 11 datasets in Fig.~\ref{fig:results11}. The AUCs for the RMSEs are shown in Fig.~\ref{fig:AUC-RMSE}, where for each dataset, we used the AUC of \texttt{BL} to normalize the AUCs of the other eight algorithms, so the AUC of \texttt{BL} was always 1. For the RMSE, a smaller AUC indicates a better performance. Similarly, we also show the AUCs of the CCs in Fig.~\ref{fig:AUC-CC}, where a larger AUC indicates a better performance. Observe that:
\begin{enumerate}
\item \texttt{RD} achieved smaller AUCs for the RMSE than \texttt{BL} on 10 of the 11 datasets, and larger  AUCs for the CC than \texttt{BL} on 9 of the 11 datasets, suggesting that \texttt{RD} is indeed effective.
\item Among the four existing ALR approaches, \texttt{GS} achieved the best average performance on both RMSE and CC. The reason can be explained as follows. In pool-based ALR we compute the RMSE and CC on all remaining unlabeled samples, and a large error on a single sample may significantly deteriorate the overall performance, i.e., the samples make unequal contributions to the RMSE and CC. A diverse sample, which is far away from currently selected samples, is more likely to give such a large error (its neighborhood has not been sufficiently modeled). \texttt{GS} considers only the diversity, and makes sure the selected samples are somewhat uniformly distributed in the entire input space, i.e., all neighborhoods in the input space are considered, and hence large errors are less likely to occur. This is different from ALC, in which all misclassified samples make equal contributions to the classification error, no matter how far away they are from the decision boundary.
\item Generally \texttt{RD-QBC}, \texttt{RD-EMCM} and \texttt{RD-GS} achieved the best performances among the nine algorithms.
\end{enumerate}

The ranks of the nine approaches on the 11 datasets, according to the AUCs, are shown in Table~\ref{tab:ranks}. Observe that on average, \texttt{RD-EMCM}, \texttt{RD-GS} and \texttt{RD-QBC} ranked among the top three on both RMSE and CC, \texttt{RD} and \texttt{GS} ranked the next, \texttt{EEMCM} slightly outperformed \texttt{EMCM}, and \texttt{BL} was the last. This again confirms the superiority of our proposed approaches.

\begin{table}[h] \centering \setlength{\tabcolsep}{.8mm}
\caption{Ranks of the nine approaches on the 11 datasets in transductive learning.}   \label{tab:ranks}
\begin{tabular}{c|l|ccccccccc}   \hline
   &     &     &     &      &    &    &  & \texttt{RD-} & \texttt{RD-}  & \texttt{RD-} \\
   &Dataset & \texttt{BL}  & \texttt{QBC} & \texttt{EMCM} & \texttt{EEMCM} & \texttt{GS} & \texttt{RD} & \texttt{QBC} & \texttt{EMCM} & \texttt{GS} \\ \hline
   &Concrete-CS        &9&8&7&6&5&1&4&2&3\\
   &IADS-Arousal       &9&8&7&6&1&5&4&2&3\\
   &Yacht              &5&8&7&3&9&1&2&6&4\\
   &autoMPG            &9&7&8&5&1&6&4&2&3\\
RMSE&NO2               &9&5&7&6&3&8&4&1&2\\
   &Housing            &9&7&6&5&8&3&2&1&4\\
   &CPS                &8&7&6&9&1&5&3&4&2\\
   &Concrete           &9&8&7&6&4&5&3&1&2\\
   &Airfoil            &9&3&7&6&8&1&2&4&5\\
   &Wine-red           &9&4&6&7&5&8&2&1&3\\
   &Wine-white         &8&4&3&7&1&9&6&2&5 \\
   &\textbf{Average}    &\textbf{9}&\textbf{7}&\textbf{8}&\textbf{6}&\textbf{4}&\textbf{5}&\textbf{2}&\textbf{1}&\textbf{2}\\  \hline
   &Concrete-CS        &9&8&7&6&5&1&4&2&3\\
   &IAPS-Arousal       &9&5&6&7&8&4&3&1&2\\
   &Yacht              &9&8&7&3&6&1&5&4&2\\
   &autoMPG            &9&6&7&5&1&8&4&2&3\\
CC &NO2                &8&5&7&4&6&9&3&2&1\\
   &Housing            &9&8&7&6&5&4&3&1&2\\
   &CPS                &8&6&7&9&1&5&2&3&4\\
   &Concrete           &9&6&7&8&5&4&3&2&1\\
   &Airfoil            &9&5&7&6&8&1&2&3&4\\
   &Wine-red           &4&7&9&6&8&5&2&1&3\\
   &Wine-white         &2&1&3&9&6&8&7&5&4  \\
   &\textbf{Average}   &\textbf{9}&\textbf{6}&\textbf{8}&\textbf{7}&\textbf{5}&\textbf{4}&\textbf{3}&\textbf{1}&\textbf{2}\\ \hline
\end{tabular}
\end{table}

\subsection{Statistical Analysis}

To determine if the differences between different pairs of algorithms were statistically significant, we also performed non-parametric multiple comparison tests on the AUCs using Dunn's procedure \cite{Dunn1961,Dunn1964}, with a $p$-value correction using the False Discovery Rate method \cite{Benjamini1995}. The $p$-values for the AUCs of RMSEs and CCs are shown in Table~\ref{tab:Dunn}, where the statistically significant ones are marked in bold. Observe that:
\begin{enumerate}
\item All ALR approaches had statistically significantly better RMSEs and CCs than \texttt{BL}.
\item Among the four existing ALR approaches, \texttt{GS} had statistically significantly better RMSEs than \texttt{QBC}, \texttt{EMCM} and \texttt{EEMCM}.
\item \texttt{RD} had statistically significantly better RMSE and CC than \texttt{QBC}, \texttt{EMCM} and \texttt{EEMCM}.
\item \texttt{RD-QBC}, \texttt{RD-EMCM} and \texttt{RD-GS} all had statistically significantly better RMSE and CC than the other six approaches, suggesting again that \texttt{RD} is complementary to \texttt{QBC}, \texttt{EMCM} and \texttt{GS}, and hence integrating \texttt{RD} with any of the latter three can further improve the performance.
\item There were no statistically significant differences among \texttt{RD-QBC}, \texttt{RD-EMCM} and \texttt{RD-GS}.
\end{enumerate}

\begin{table}[htpb] \centering \setlength{\tabcolsep}{.4mm}
\caption{$p$-values of non-parametric multiple comparisons on the AUCs of RMSEs and CCs in transductive learning.}   \label{tab:Dunn}
\begin{tabular}{c|l|cccccccc}   \hline
& &  &           &       &    &   & & RD-  &  RD- \\
&   & BL &           QBC &     EMCM  & EEMCM &  GS &  RD & QBC  &  EMCM \\ \hline
&QBC & \textbf{.0000} & & & &&&&\\
&EMCM & \textbf{.0000} & .4662 & & & & &&\\
&EEMCM & \textbf{.0000} & .1399 & .1252& & & &\\
&GS & \textbf{.0000} & \textbf{.0000} & \textbf{.0000} &\textbf{.0019} & & &&\\
RMSE&RD & \textbf{.0000} & \textbf{.0000} & \textbf{.0000} & \textbf{.0002} &.2585 & && \\
&RD-QBC &\textbf{.0000} & \textbf{.0000} & \textbf{.0000} & \textbf{.0000} & \textbf{.0000} &\textbf{.0004} & &\\
&RD-EMCM & \textbf{.0000} & \textbf{.0000} & \textbf{.0000} & \textbf{.0000} &\textbf{.0000} & \textbf{.0000}
&.2115&\\
&RD-GS & \textbf{.0000} & \textbf{.0000} & \textbf{.0000} & \textbf{.0000} & \textbf{.0001} & \textbf{.0014}
& .3512 &.1219\\   \hline
&QBC & \textbf{.0000} & & & &&&&\\
&EMCM & \textbf{.0000} & .0654 & & & & &&\\
&EEMCM & \textbf{.0000} & \textbf{.0006} &.0499 & & & &&\\
&GS & \textbf{.0000} & .4417 & .0499 & \textbf{.0004} & & &&\\
CC&RD & \textbf{.0000} & \textbf{.0217} & \textbf{.0002} & \textbf{.0000} &.0299 & & &\\
&RD-QBC & \textbf{.0000} & \textbf{.0000} & \textbf{.0000} & \textbf{.0000} & \textbf{.0000} & \textbf{.0004}& &\\
&RD-EMCM & \textbf{.0000} & \textbf{.0000} & \textbf{.0000} & \textbf{.0000} &\textbf{.0000} & \textbf{.0000}
&.1847&\\
&RD-GS & \textbf{.0000} & \textbf{.0000} & \textbf{.0000} & \textbf{.0000} & \textbf{.0000} & \textbf{.0000}
& .2367 &.4340\\   \hline
\end{tabular}
\end{table}

\subsection{Visualization}

It's also interesting to visualize the sample selection results of different algorithms to confirm the superiority of the RD based ALR approaches. However, because the feature spaces had at least seven dimensions, it is difficult to visualize them directly. So, we performed  principle component analysis (PCA) on each dataset, and represented all samples as their projections on the first two principle components. Due to the page limit, we only show the results for the Concrete-CS dataset (more results can be found in the Supplementary Materials). The red asterisks in Fig.~\ref{fig:Concrete-CS-PCA0} indicate the initial seven selected samples. Observe that random initialization, which was used in \texttt{BL}, \texttt{QBC}, \texttt{EMCM} and \texttt{GS}, may leave a large portion of the feature space unsampled. However, \texttt{RD}, \texttt{RD-QBC}, \texttt{RD-EMCM} and \texttt{RD-GS}, which used our proposed initialization approach, initialized the samples more uniformly in the entire feature space. The red asterisks in Fig.~\ref{fig:Concrete-CS-PCA} indicate the final 20 samples selected by different algorithms. Observe that:
\begin{enumerate}
\item \texttt{BL} still left a large region of the feature space unsampled, even after 20 samples.
\item \texttt{QBC}, \texttt{EMCM} and \texttt{GS} selected one or more samples near the boundary of the feature space, which may be outliers. On the contrary, \texttt{EEMCM} and \texttt{RD} selected samples uniformly distributed in the whole feature space, and no selected samples were obvious outliers.
\item The samples selected by \texttt{RD-QBC} distributed in the feature space more uniformly than those selected by \texttt{QBC}. Similar patterns can also be observed between \texttt{RD-EMCM} and \texttt{EMCM}, and between \texttt{RD-GS} and \texttt{GS}.
\end{enumerate}
In summary, the PCA visualization results confirm that the RD based ALR approaches selected more reasonable samples, which resulted in better regression performances.

\begin{figure}[htpb]\centering
\subfigure[]{\label{fig:Concrete-CS-PCA0}     \includegraphics[width=\linewidth,clip]{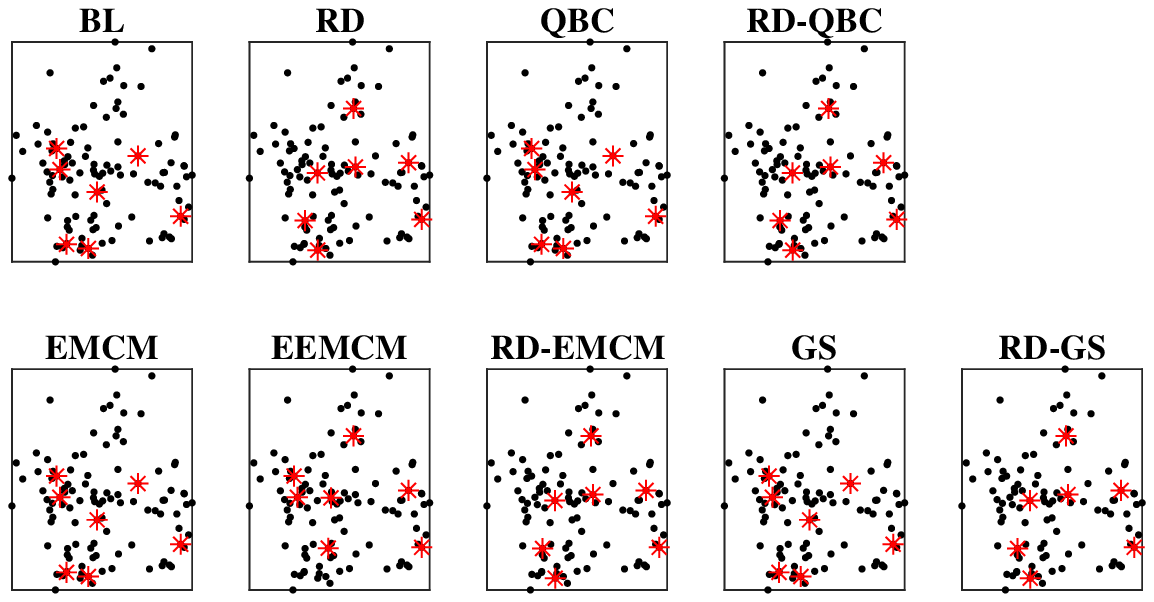}}
\subfigure[]{\label{fig:Concrete-CS-PCA}     \includegraphics[width=\linewidth,clip]{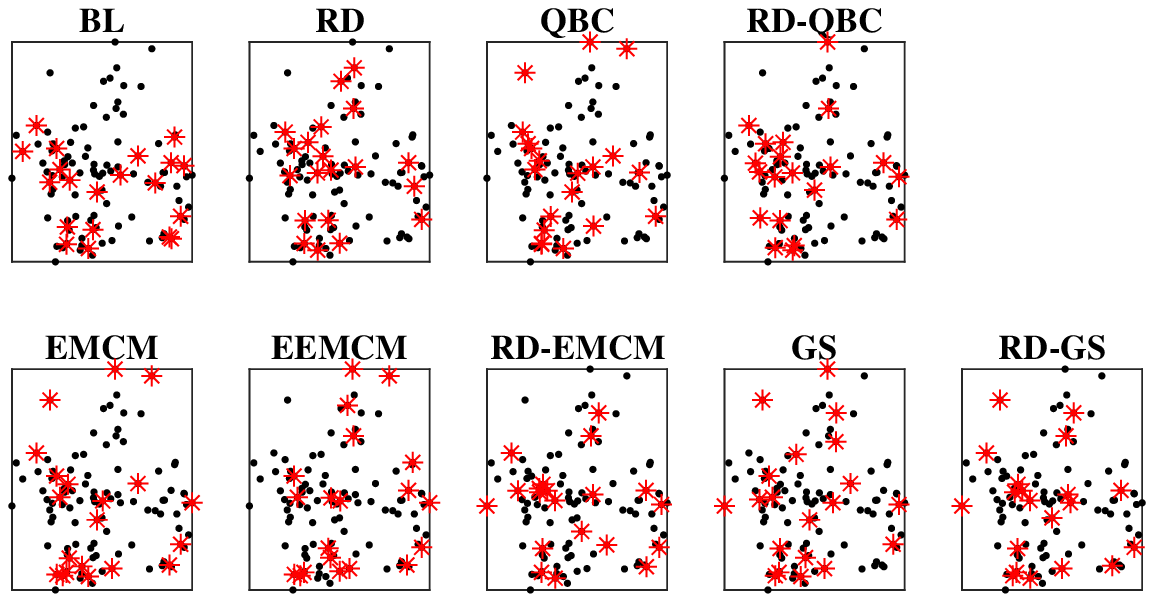}}
\caption{PCA visualization of the selected samples (red asterisks) by different algorithms on the Concrete-CS dataset. (a) the initial seven selected samples; (b) the final 20 selected samples.} \label{fig:PCA}
\end{figure}

\subsection{Individual Improvements}

Table~\ref{tab:ranks} shows that \texttt{RD-EMCM} achieved the best average RMSE among the nine algorithms. Recall that \texttt{RD-EMCM} has three enhancements over the random sampling approach (\texttt{BL}):
\begin{enumerate}
\item \textit{Enhancement 1}: \texttt{RD-EMCM} considers both the representativeness and the diversity in initializing the first $d$ samples, but \texttt{BL} does not consider either of them.
\item \textit{Enhancement 2}: \texttt{RD-EMCM} considers both the representativeness and the diversity in selecting the new sample in each iteration, but \texttt{BL} does not consider either of them.
\item \textit{Enhancement 3}: \texttt{RD-EMCM} considers also the informativeness in selecting the new sample in each iteration, but \texttt{BL} does not consider it.
\end{enumerate}
It is interesting to study if each of the three enhancements is necessary, and if so, what their individual effect is.

\begin{figure}[htpb]\centering
\subfigure[]{\label{fig:AUC-RMSE3}     \includegraphics[width=.88\linewidth,clip]{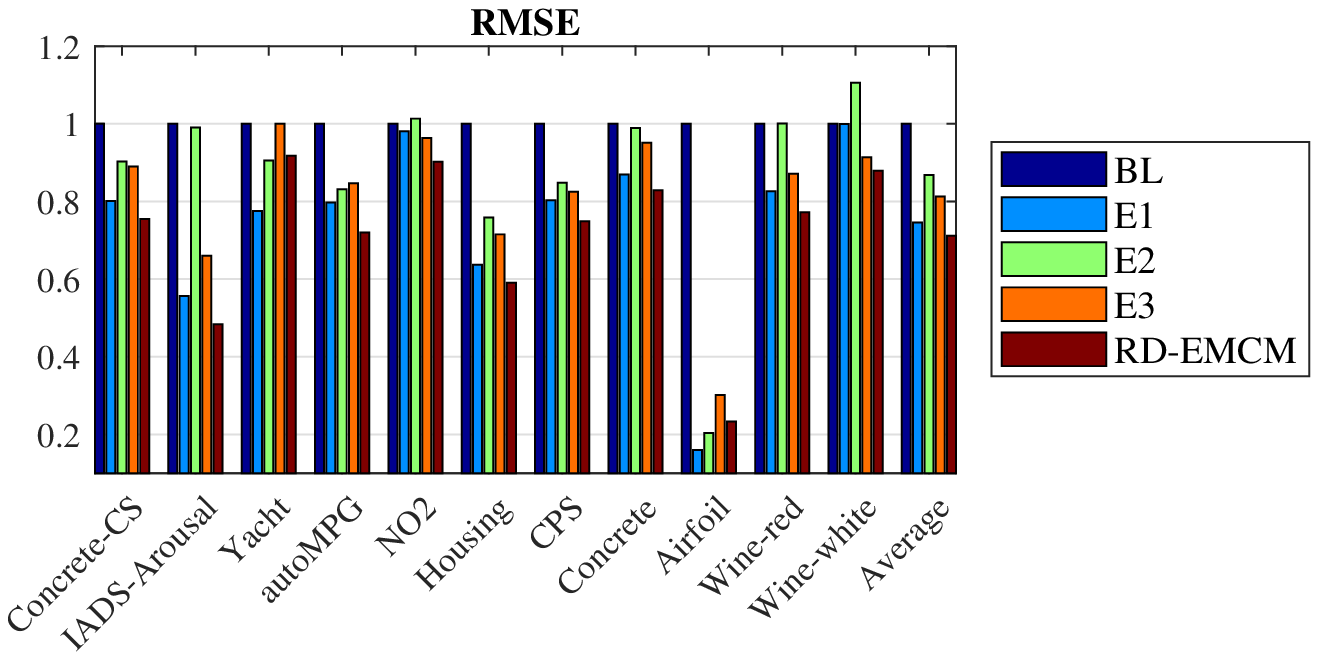}}
\subfigure[]{\label{fig:AUC-CC3}     \includegraphics[width=.88\linewidth,clip]{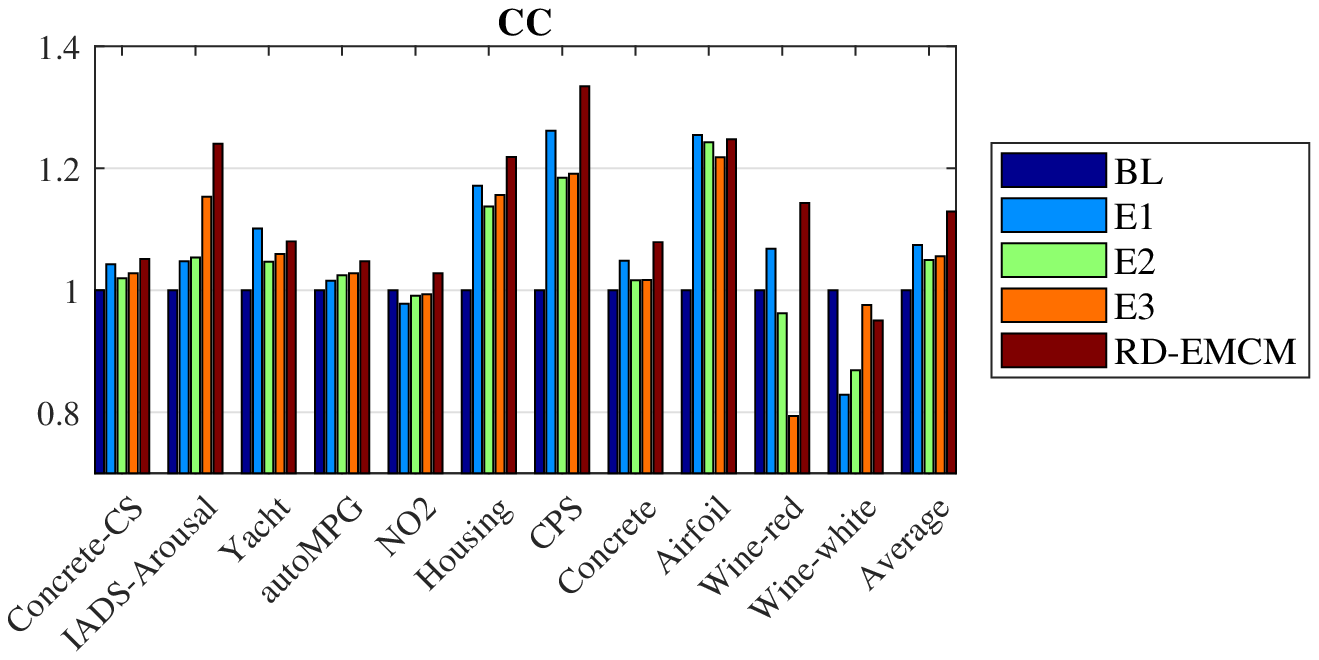}}
\caption{AUCs of the five algorithms over the 11 datasets, averaged over 30 runs. (a) RMSE; (b) CC.} \label{fig:AUC3}
\end{figure}

For this purpose, we constructed three modified versions of \texttt{RD-EMCM}, by considering each enhancement individually: \texttt{E1}, which employs only the first enhancement on more representative and diverse initialization; \texttt{E2}, which employs only the second enhancement on more representative and diverse sampling in each iteration; and, \texttt{E3}, which employs only the third enhancement on more informative sampling in each iteration. We then compared their performances with \texttt{BL} and \texttt{RD-EMCM}. Due to the page limit, we only show the results on the CPS dataset (averaged over 30 runs) in Fig.~\ref{fig:CPS3} (more results can be found in the Supplementary Materials). The AUCs for RMSEs and CCs for all 11 datasets are shown in Fig.~\ref{fig:AUC3}.

Observe that all three enhancements outperformed \texttt{BL} on most datasets, especially for the RMSE, which was directly optimized in the objective function of ridge regression. More specifically, Fig.~\ref{fig:CPS3} shows that the first enhancement on more representative and diverse initialization helped when $m$ was very small; the second and third enhancements helped when $m$ became larger. By combining the three enhancements, \texttt{RD-EMCM} achieved the best performance at both small and large $m$. This suggests that the three enhancements are complementary, and they are all essential to the improved performance of \texttt{RD-EMCM}.

\begin{figure}[htpb] \centering
\includegraphics[width=.7\linewidth,clip]{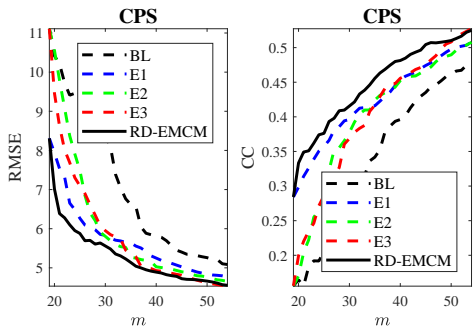}
\caption{Effect of the individual enhancements on the CPS dataset.} \label{fig:CPS3}
\end{figure}

\subsection{Inductive Learning Results} \label{sect:In}

The results presented in this section so far focused only on transductive learning. This subsection presents the inductive learning results, i.e., testing results on the 20\% samples in $\mathbf{P}$ but not in $\mathbf{P}_{80}$. The AUCs of the nine algorithms on the 11 datasets are shown in Fig.~\ref{fig:AUC-In} (more results can be found in the Supplementary Materials). Observe that Fig.~\ref{fig:AUC-In} is very similar to Fig.~\ref{fig:AUC}. Our conclusions drawn in transductive learning still hold in inductive learning.

\begin{figure}[htpb]\centering
\subfigure[]{\label{fig:AUC-RMSE-In}     \includegraphics[width=.88\linewidth,clip]{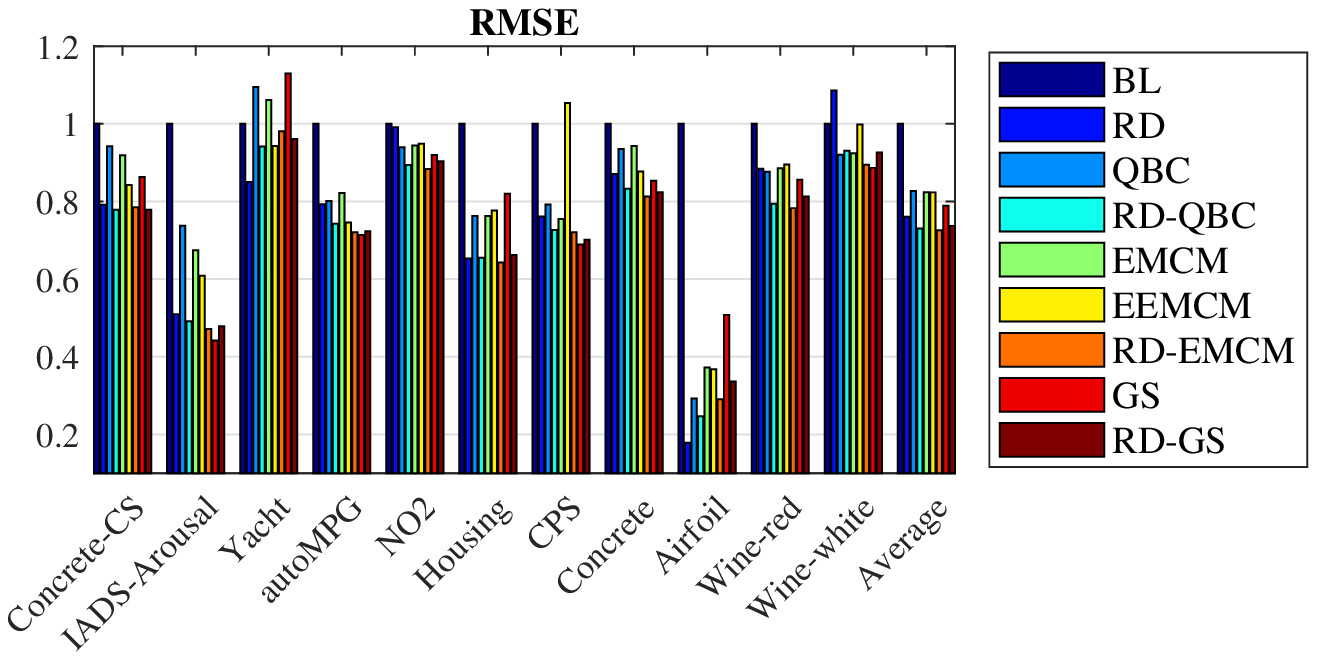}}
\subfigure[]{\label{fig:AUC-CC-In}     \includegraphics[width=.88\linewidth,clip]{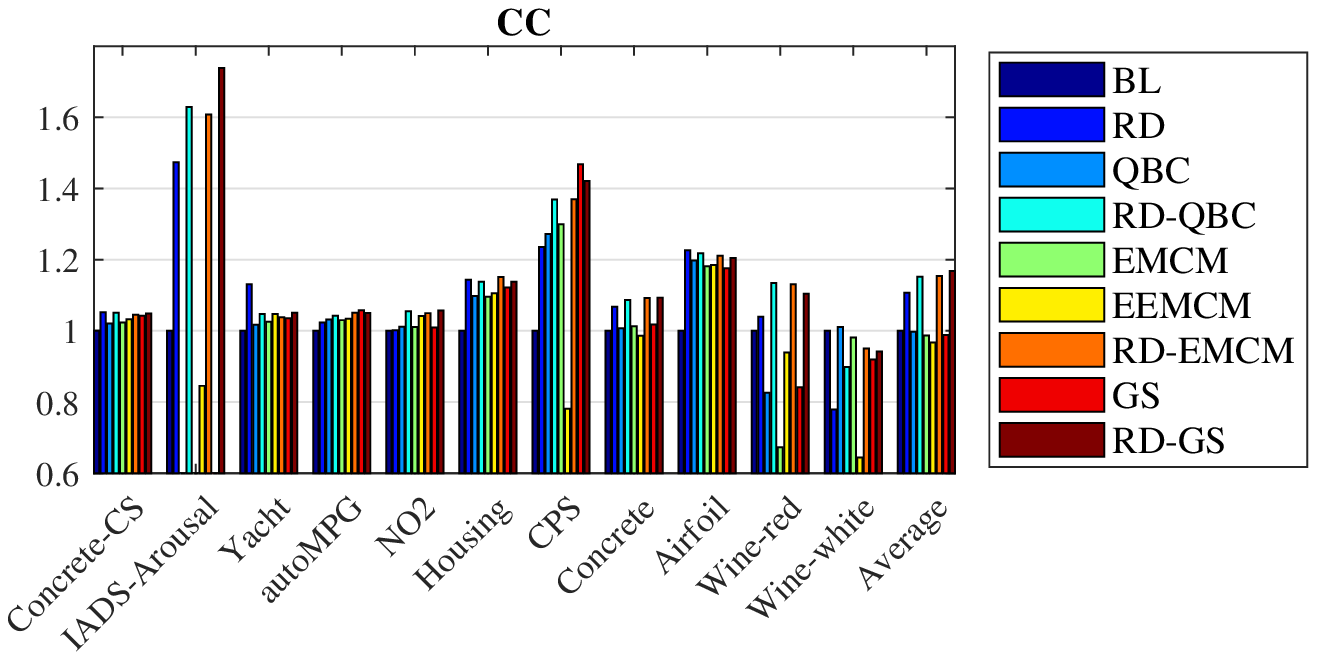}}
\caption{AUCs of the nine algorithms on the 11 datasets in inductive learning, averaged over 100 runs. (a) RMSE; (b) CC.} \label{fig:AUC-In}
\end{figure}

\section{Conclusions} \label{sect:conclusions}

AL has been frequently used to reduce the data labeling effort in machine learning. However, most existing AL approaches are for classification. This paper studied AL for regression. We proposed three essential criteria that should be considered in selecting a new sample in pool-based sequential ALR, which are informativeness, representativeness, and diversity. An ALR approach called RD was proposed, which considers both the representativeness and diversity in both initialization and subsequent iterations. The RD approach can also be integrated with existing pool-based sequential ALR approaches, such as QBC, EMCM and GS, to further improve the performance. Extensive experiments on 11 public datasets from various domains confirmed the effectiveness of our proposed approaches.

%\bibliographystyle{IEEETranS}\bibliography{drwubib}
% Generated by IEEEtranS.bst, version: 1.14 (2015/08/26)

\end{document}